\documentclass[11pt]{article}
\usepackage{geometry}
\usepackage{coling2020}
\usepackage{times}
\usepackage{url}
\usepackage{latexsym}
\usepackage{microtype}
\usepackage{amsmath}
\usepackage{amssymb}
\usepackage{siunitx}
\usepackage{booktabs}
\usepackage{etoolbox}
\usepackage{graphics}
\usepackage{graphicx}
\usepackage{caption}
\usepackage{subcaption}
\usepackage{floatrow}

\DeclareCaptionLabelFormat{andtable}{#1~#2  \&  \tablename~\thetable}

\colingfinalcopy 

\usepackage{xcolor}
\usepackage{soul}

\setstcolor{red}

\title{IDS at SemEval-2020 Task 10: Does Pre-trained Language Model Know What to Emphasize?}

\author{Jaeyoul Shin, Taeuk Kim and Sang-goo Lee \\
  Dept. of Computer Science and Engineering\\
  Seoul National University, Seoul, Korea \\
  {\tt \{jyshin,taeuk,sglee\}@europa.snu.ac.kr}}

\date{}

\begin{document}
\maketitle
\begin{abstract}
    We propose a novel method that enables us to determine words that deserve to be emphasized from written text in visual media, relying only on the information from the self-attention distributions of pre-trained language models (PLMs).
    With extensive experiments and analyses, we show that 1) our zero-shot approach is superior to a reasonable baseline that adopts TF-IDF and that 2) there exist several attention heads in PLMs specialized for emphasis selection, confirming that PLMs are capable of recognizing important words in sentences.
\end{abstract}

\section{Introduction}
\label{intro}

%
%
\blfootnote{
    %
    %
    %
    %
    %
    %
    \hspace{-0.65cm}  
    This work is licensed under a Creative Commons 
    Attribution 4.0 International License.
    License details:
    \url{http://creativecommons.org/licenses/by/4.0/}.
}

In terms of visual communication such as social media posts (posters, flyers, and ads) and motivational messages, text emphasis is crucial in that it facilitates the comprehension of written text and that it helps convey the author’s intent.
Therefore, it is expected that the development of an automatic system that recommends which part to emphasize in the text of visual media will bring significant advantages; e.g., it is possible to accelerate the making process of posters and videos for advertisement.
In this paper, we attempt to devise such a system solely with the aid of pre-trained language models (PLMs) without any task-specific laborious training.

Recently, there has been a substantial amount of work in the literature to figure out what knowledge Transformer \cite{vaswani2017attention} based PLMs such as BERT \cite{devlin2019bert} contain and why they perform surprisingly well on various downstream tasks \cite{goldberg2019assessing,kovaleva2019revealing,rogers2020primer}.
Among these, a group of studies has focused on analyzing PLMs' self-attention distributions to find some evidence that supports the existence of linguistic knowledge within the pre-trained weights of PLMs.
Specifically, \newcite{clark2019does} investigated BERT's individual attention heads to probe its ability to parse dependency trees, while \newcite{Kim2020Are} proposed an unsupervised constituency parsing method applicable on top of the distributions.

Following the same philosophy shared by the work mentioned above, we propose a zero-shot emphasis selection method, assuming that during pre-training, some attention heads in PLMs are inspired to recognize which words are more important than others.
We test our method on a carefully designed dataset \cite{shirani2020semeval}, and it records the ranking score 0.690 on the validation set, which outperforms an intuitive baseline adopting the term frequency-inverse document frequency (TF-IDF) strategy \cite{jones1972statistical}.
Furthermore, our method operates in a fully zero-shot manner, not leveraging gold-standard annotations provided by the dataset at all, implying its universal applicability in a low/zero-resource regime.

\section{Task}
This paper aims to present a tractable solution to the SemEval 2020 shared task 10 \cite{shirani2020semeval}.
The dataset provided by this task consists of short English sentences obtained from Adobe Spark. It contains a variety of subjects featured in flyers, posters, and advertisements or motivational memes on social media---for example, ``In honor of the brave'' \cite{shirani2019learning}.

\newcommand{\comment}[1]{}

\newfloatcommand{capbtabbox}{table}[][\FBwidth]
\begin{figure}[t!]
\begin{floatrow}\CenterFloatBoxes
    \capbtabbox[0.65\textwidth]{%
        \scriptsize
        \sisetup{detect-weight,mode=text}
        \renewrobustcmd{\bfseries}{\fontseries{b}\selectfont}
        \renewrobustcmd{\boldmath}{}
        \newrobustcmd{\B}{\bfseries}
        \begin{tabular}{c c c c c c c c c c c} \hline
            \B Words & \B A1 & \B A2 & \B A3 & \B A4 & \B A5 & \B A6 & \B A7 & \B A8 & \B A9 & \B $e\_freq$ \\ \hline
            In & O & I & O & O & O & O & O & O & I & 0.222 \\
            honor & I & I & O & O & I & I & I & I & I & 0.778 \\
            of & O & O & O & O & O & O & O & O & I & 0.111 \\
            the & O & O & O & O & O & O & O & O & I & 0.111 \\
            brave & O & I & I & I & O & I & I & I & I & 0.778 \\ \hline
        \end{tabular}    
    }{%
        \caption{Data structure of the example sentence ``In honor of the brave''. A1-A9 are nine annotators. I and O correspond to whether to emphasize the word or not. Emphasis frequency, $e\_freq$, is the average of nine labels of A1-A9.}%
    }
    \ffigbox[.30\textwidth]{%
        \scriptsize
        \includegraphics[width=50mm]{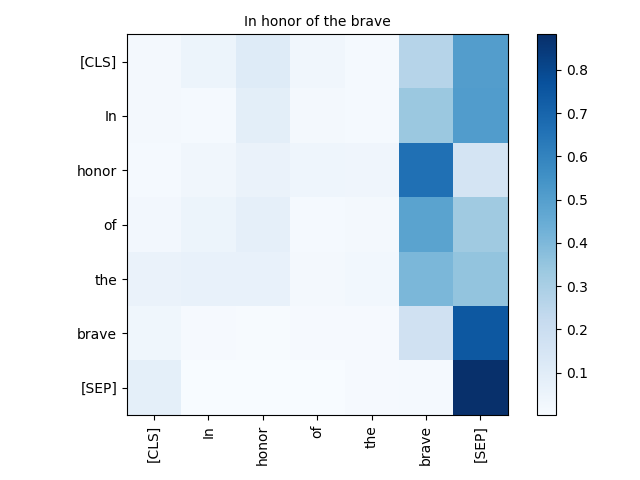}
    }{
        \caption{Sample attention map of the example sentence.}
    }
\end{floatrow}
\end{figure}

\comment{
{\centering
\begin{minipage}{0.55\linewidth}
    \centering
    \scriptsize
    \sisetup{detect-weight,mode=text}
    \renewrobustcmd{\bfseries}{\fontseries{b}\selectfont}
    \renewrobustcmd{\boldmath}{}
    \newrobustcmd{\B}{\bfseries}
    \begin{tabular}{c c c c c c c c c c c} \hline
        \B Words & \B A1 & \B A2 & \B A3 & \B A4 & \B A5 & \B A6 & \B A7 & \B A8 & \B A9 & \B $e\_freq$ \\ \hline
        In & O & I & O & O & O & O & O & O & I & 0.222 \\
        honor & I & I & O & O & I & I & I & I & I & 0.778 \\
        of & O & O & O & O & O & O & O & O & I & 0.111 \\
        the & O & O & O & O & O & O & O & O & I & 0.111 \\
        brave & O & I & I & I & O & I & I & I & I & 0.778 \\ \hline
    \end{tabular}
   \captionof{table}{Data structure of the example sentence ``In honor of the brave''. A1-A9 are nine annotators. I and O each corresponds to whether emphasize or not. Emphasis frequency, $e\_freq$, is the average of nine labels of A1-A9.}
\end{minipage}
\hfill
\begin{minipage}{0.40\linewidth}
   \centering
   \scriptsize
   \includegraphics[width=50mm]{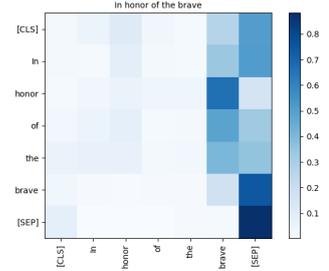}
   \captionof{figure}{Sample attention map of the example sentence. \\}
\end{minipage}
}
}

In detail, as shown in Table 1, each word of a sentence in the dataset is provided with nine binary labels (I/O tags) that correspond to the decisions of nine annotators, each indicating whether to emphasize the target word or not.
Here we define the emphasis frequency ($e\_freq$) of the word in the sentence as the average of these nine labels, where treating `I' as 1 and `O' as 0.
Our goal is to construct a model that predicts the correct ranking of words based on each word's gold emphasis frequency.

For evaluation, we use the \(Match_{m}\) $(m \in [1,2,3,4])$ score which is calculated as:
\begin{equation}
Match_{m} = \frac{\sum_{x \in D} |S_{m}^{(x)} \cap \hat{S}_{m}^{(x)} | / m}{|D|},
\end{equation}
where \(S_{m}^{(x)}\) is a set of top-$m$ high emphasis frequency words in an input sentence $x$ of a dataset $D$.; e.g., \(S_{2}^{(``In \: honor \: of \: the \: brave'')}\) would be ['honor', 'brave'].
\(\hat{S}_{m}^{(x)}\) is a set of top-$m$ high emphasis frequency words based on our model's prediction.
$|\cdot|$ corresponds to the number of elements in a set.
Moreover, we introduce the \(Ranking\_Score\) as an aggregated measure that averages all possible \(Match_{m}\) scores:
\begin{equation}
Ranking\;Score = \frac{\sum_{m \in [1,2,3,4]} Match_{m} }{4}.
\end{equation}

\comment{
\section{Related Work}


PLMs are usually trained on heavily large corpora and are known to be able to acquire some linguistic knowledge during pre-training. 
Many researchers have studied what kind of knowledge they contain by analyzing their last hidden states, internal representations, and attention distributions.
In our case, we focus on the attention distributions, which are attention weights computed by PLMs' attention heads, investigating whether they can be utilized as features for computing each word's emphasis frequency.



Understanding the inner workings of the attention heads of PLMs is fundamental to analyze Transformer \cite{vaswani2017attention} based language models.
\newcite{clark2019does} probed individual attention heads of BERT 
by evaluating the performance of each attention head on dependency parsing.
Similarly, \newcite{Kim2020Are} proposed an unsupervised constituency parsing method applicable on the attention distributions of various PLMs.
Both studies suggest that there exist a set of specialized attention heads in PLMs that are sensitive to some syntactic structures of input sentences and therefore can be utilized as features for syntactic tasks. 

Inspired by these observations, we also attempt to discover another group of meaningful attention heads of PLMs\footnote{We consider the following PLMs as candidates for our method: BERT \cite{devlin2019bert}, DistilBERT \cite{sanh2019distilbert}, GPT-2 \cite{radford2018improving}, RoBERTa \cite{liu2019roberta}, XLNet \cite{yang2019xlnet}, and XLM \cite{conneau2019cross}. 
Note that we leverage both English and multilingual PLMs. 
} that are able to recognize some semantic or functional keywords in a sentence.
}

\section{Method}

In this section, we propose three ways to induce the emphasis frequency (\(e\_freq\)) of each word in a sentence using a PLM's\footnote{We consider the following PLMs as candidates for our method: BERT \cite{devlin2019bert}, DistilBERT \cite{sanh2019distilbert}, GPT-2 \cite{radford2018improving}, RoBERTa \cite{liu2019roberta}, XLNet \cite{yang2019xlnet}, and XLM \cite{conneau2019cross}} attention maps.
The intuition behind our approach is that the more one word draws attention from other words, the more suitable this word as a target to be emphasized.
In other words, the words contribute the most to construct the intermediate representations of other words for the next layer of PLMs should have high \(e\_freq\) values.
As we only resort to the inherent knowledge of PLMs rather than learning a separate model from scratch, we do not need any further training based on supervision from gold-standard annotations to implement our approach.
This characteristic is attractive in a perspective that the annotations required to build gold-standard labels are too expensive and even somewhat subjective.

To reduce the ambiguity, here we define terminology.
We denote a sentence as a set of words, $s = \{w_m | m = 1, \dots , n\}$, where $n$ stands for the number of words in the sentence.
When a sentence $s$ is fed to a PLM, an attention map, which is a set of attention distributions of a particular self-attention head, can be extracted.
We define $G$ as a set of attention maps extracted from a PLM; i.e., $G = \{g_{(i,j)} \in \mathbb{R}^{\emph{(n+2)} \times \emph{(n+2)}} | i=1, \dots , l, j = 1, \dots, a \}$, where $g_{(i,j)}$ is an attention map of the $j$th attention head on the $i$th layer, and $l$ and $a$ are the numbers of layers and attention heads per layer, respectively.
The reason why we add $2$ to $n$ is to consider two special tokens, [CLS] and [SEP].
In Figure 1, there is a sample attention map of the example sentence, where each row represents an attention distribution of the corresponding word; 
e.g., The $3^{rd}$ row is an attention distribution of the word 'honor' over other words.

There are some pre-processing procedures to obtain proper attention maps which can be utilized for our methods.
First, we add special tokens [CLS] and [SEP] to an input sentence.---for example, ``[CLS] In honor of the brave [SEP]'' as described in \newcite{devlin2019bert}.
Then, we can extract attention maps after feeding an input sentence to a PLM.
Since most PLMs tokenize a sentence into subword-level, we convert token-level attention maps to word-level attention maps by averaging each group of the attention weights of subword tokens that belong to the same word following \newcite{clark2019does} and \newcite{Kim2020Are}.

For each attention head, we consider three options to derive \(e\_freq(word)\): Words2Target, CLS2Target and SEP2Target.
Each option is depicted in Figure 2.

\begin{figure}[!t]
\small
  \centering
  \begin{minipage}[b]{0.1\linewidth}
  \end{minipage}
  \hfill
  \begin{minipage}[b]{0.25\linewidth}
    \includegraphics[width=\linewidth]{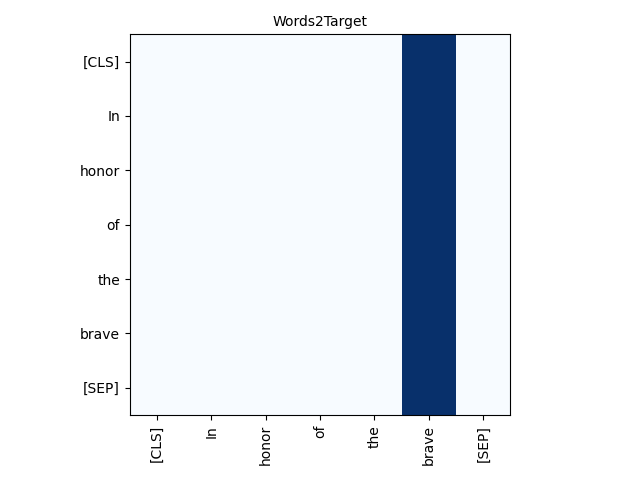}
    \subcaption{Words2Target}
  \end{minipage}
  \hfill
  \begin{minipage}[b]{0.25\linewidth}
    \includegraphics[width=\linewidth]{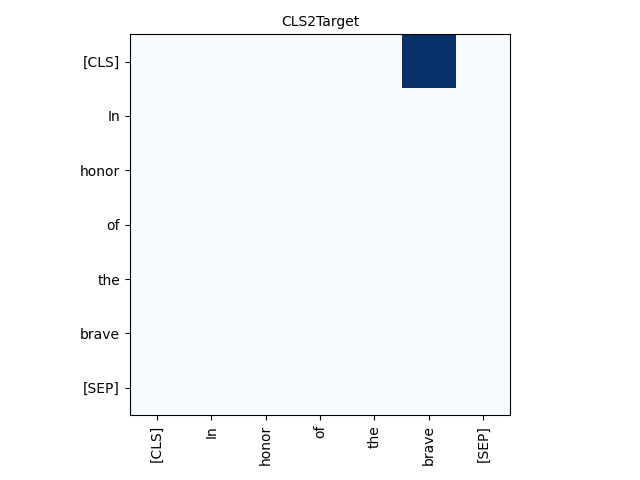}
    \subcaption{CLS2Target}
  \end{minipage}
  \hfill
    \begin{minipage}[b]{0.25\linewidth}
    \includegraphics[width=\linewidth]{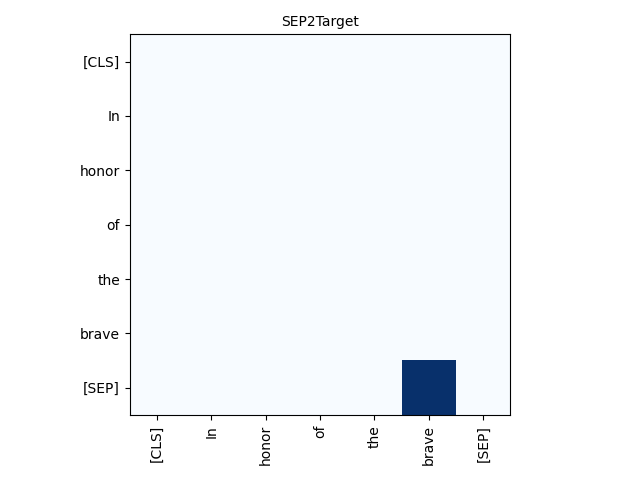}
    \subcaption{SEP2Target}
  \end{minipage}
  \hfill
  \begin{minipage}[b]{0.1\linewidth}
  \end{minipage}
  \caption{Three ways to derive emphasis frequency, $e\_freq$, of a target word 'brave'. Dark areas are where each method refers to.}
\end{figure}

\subsection{Words2Target}
Given a particular attention head (the $j$th attention head on the $i$th layer extracted from a PLM), the emphasis frequency of a word can be calculated as an average over other words' attention weights on the target word:
\begin{equation}
    e\_freq(word_{t})_{Words2Target} = \frac{\sum_{k=1}^{n+2} g_{(i,j)}[k][t]}{n+2}
\end{equation}
This equation has a meaning of how much the $t$th word, $word_{t}$, is influential when constructing the hidden representations of other words.
It also can be thought of as the average of values on the $t$th column of the attention map as shown in Figure 2 - (a).

\subsection{CLS2Target and SEP2Target}
Including BERT, many PLMs use special tokens ([CLS], [SEP]) to encode a sentence representation or the relation of two input sentences.
The attention weight of a special token on \(word_{t}\) means how much the \(word_{t}\) contributes to sentence representation.
Thus, when a PLM is given an input sequence ([CLS], $w_1$, $w_2$, $\dots$, $w_n$, [SEP]), we induce\footnote{These methods can't be applied to GPT-2 model because it doesn't use special tokens.} \(e\_freq(word_{t})\) from the attention weight of \(word_{t}\) on both special tokens, as expressed in Figure 2 - (b), (c):
\begin{equation}
    e\_freq(word_{t})_{CLS2Target} =  g_{(i,j)}[idx_{[CLS]}][t],
\end{equation}
\begin{equation}
    e\_freq(word_{t})_{SEP2Target} =  g_{(i,j)}[idx_{[SEP]}][t].
\end{equation}

\subsection{Best Configuration Selection}
For a single PLM, there are \(l \times a \times 3 \) possible configurations
because \(e\_freq(word_{t})\) can be computed in three ways of $M \in [Words2Target, CLS2Target, SEP2Target]$ for all $g_{(i,j)} \in G$.
For instance, in the case of the BERT-base, which consists of 12 layers and 12 attention heads per layer, there are 12 $\times$ 12 $\times$ 3 $=$ 432 possible configurations.
We compute \(Match_{m}\) and $Ranking\;Score$ scores for every $\left( g_{(i,j)} , M \right)$ pair and select the best configuration for the PLM based on the $Ranking\;Score$.

\subsection{Baseline}
To evaluate the performance of our method more precisely, we propose a reasonable baseline using Term Frequency-Inverse Document Frequency (TF-IDF) as follows:
\begin{equation}
    e\_freq(word_{t}, d_{dev})_{TF-IDF} =  \frac{f_{word_t,d_{dev}}}{len(d_{dev})} \times log \frac{|D_{train}|}{|d_{train} \in D_{train} : word_{t} \in d_{train}|+1}
\end{equation}
where $D_{train}$ and $D_{dev}$ are sets of sentences of the training set and validation set, and $d_{train}$ and $d_{dev}$ are sentences sampled respectively from these datasets.
$f_{word_t,d_{dev}}$ means the number of occurrences of $word_t$ in the sentence $d_{dev}$.
$len(d_{dev})$ is the number of words in $d_{dev}$.
The more $word_{t}$ occurs in the sentence and less included in the whole training set, the larger TF-IDF for $word_t$ in the sentence will be.
This is why the TF-IDF is considered a statistical measure of word specialty in a particular document.
Word counting method assigns $\frac{1}{f_{word_t,D_{train}}}$ which leads to rare words having larger $e\_freq$ values.
In addition, the random baseline method randomly gives $e\_freq$ value of the target word.
In experiments, we show that the TF-IDF model performs  better than the other two methods, making it a suitable option as a reasonable baseline.

\begin{table}[!t]
\centering
\scriptsize
\label{tab1}
\sisetup{detect-weight,mode=text}
\renewrobustcmd{\bfseries}{\fontseries{b}\selectfont}
\renewrobustcmd{\boldmath}{}
\newrobustcmd{\B}{\bfseries}
\begin{tabular}{l c c c c c c c c c}
\B Model             & \B L & \B A & \B Method & \B M1 & \B M2 & 
\B M3 & \B M4 & \B R dev & \B R test\\
\hline
\B Baseline & & & & & & & & &  \\
Random & - & - & - & 0.1734 & 0.3090 & 0.3749 & 0.4522 & 0.3273 & 0.3176 \\
Word Counting & - & - & - & 0.2857 & 0.4775 & 0.5625 & 0.6295 & 0.4888 & 0.5113 \\
TF-IDF & - & - & - & 0.3061 & 0.4615 & 0.6146 & 0.6758 & 0.5145 & 0.5184 \\
\hline
\B English Pre-trained LMs & & & & & & & &   \\
BERT-base-cased & 10 & 10 & Word2Target & 0.4388 & 0.6048 & 0.7045 & 0.7394 & 0.6219 & - \\
BERT-base-uncased & 10 & 8 & Word2Target & 0.4311 & 0.6247 & 0.7254 & 0.7652 & \B 0.6366 & \B 0.6249 \\
BERT-large-cased & 11 & 4 & CLS2Target & 0.4490 & 0.6286 & 0.6932 & 0.725 & 0.6240 & - \\
BERT-large-uncased & 15 & 4 & Word2Target & 0.4490 & 0.6233 & 0.7358 & 0.7598 & \B 0.6420 & \B 0.6287 \\
DistilBERT-base-cased & 5 & 10 & Word2Target & 0.4362 & 0.6233 & 0.7083 & 0.7530 & \B 0.6302 & \B 0.6130 \\
DistilBERT-base-uncased & 5 & 8 & Word2Target & 0.4541 & 0.6194 & 0.7263 & 0.7682 & \B 0.6420 & \B 0.6288 \\
GPT-2 & 3 & 10 & Word2Target & 0.2245 & 0.4350 & 0.5691 & 0.625 & 0.4634 & - \\
RoBERTa-base & 3 & 5 & CLS2Target & 0.4413 & 0.5889 & 0.6884 & 0.7152 & 0.6084 & - \\
RoBERTa-large & 10 & 14 & Word2Target & 0.4158 & 0.5703 & 0.6619 & 0.7023 & 0.5876 & - \\
XLNet-base & 2 & 4 & Word2Target & 0.2985 & 0.4204 & 0.4848 & 0.5129 & 0.4292 & - \\
XLNet-large & 3 & 11 & Word2Target & 0.3010 & 0.4403 & 0.4953 & 0.5530 & 0.4474 & - \\
XLM-mlm & 1 & 16 & Word2Target & 0.3776 & 0.5650 & 0.6496 & 0.7068 & 0.5747 & - \\
\hline
\B Multilingual Pre-trained LMs & & & & & & & &   \\
BERT-base-multilingual & 8 & 6 & Word2Target & 0.4490 & 0.6207 & 0.7083 & 0.7409 & 0.6297 & - \\
DistilBERT-base-multilingual & 4 & 6 & Word2Target & 0.4362 & 0.6260 & 0.7140 & 0.7614 & \B 0.6344 & \B 0.6199 \\
XLM-mlm-17 & 1 & 2 & Word2Target & 0.4260 & 0.5584 & 0.6203 & 0.6674 & 0.5680 & - \\
XLM-mlm-100 & 15 & 12 & SEP2Target & 0.3673 & 0.5782 & 0.6761 & 0.7303 & 0.5880 & -  \\
\hline
\B Ensemble & - & - & - & 0.4847 & 0.6790 & 0.7803 & 0.8152 & \B 0.6898 & \B 0.6656 \\
\hline
\B Supervised Baseline & - & - & - & 0.5918 & 0.7520 & 0.8040 & 0.8220 & 0.7424 & 0.75 \\
\end{tabular}

\caption{Experimental results. L: layer number, A: attention head number, Method: Word2Target / CLS2Target / SEP2Target, M1-M4: $Match_{1}$-$Match_{4}$ scores on the validation set, R dev: Ranking score on the validation set, R test: Ranking score on the test set. Bold numbers correspond to the top 5 ranking scores of single models, and ensemble model. We report R test scores only for the top-5 models and the ensemble model.}
\end{table}

\section{Results}

In Table 2, we report the results of our method to various PLMs on the validation set and test set.
Note that without a few exceptions, our method combined with PLMs performs better than baselines, including the one with TF-IDF (0.5145 ranking score).
Among PLMs, the BERT-large-uncased and DistilBERT-base-uncased models record the best performance. 

Here we mention several takeaways from our results.
First, particular PLMs (GPT-2, XLNet) report comparably lower ranking scores against the other PLMs and the TF-IDF baseline.
Since GPT-2 adopts a Transformer decoder to its architecture, its attention distribution is leaned on the first word of a sentence \cite{vig2019visualizing}.
This results in a sub-optimal solution where each first word for all 390 sentences in the validation set becomes the top-$1$ $e\_freq$ high score word.
On the other hand, XLNet model tends to focus more on punctuation marks such as period (.) and comma (,).
Specifically, the XLNet-base model predicts that a period should be one of the top-$4$ high \(e\_freq\) words for 289 sentences in the validation set, even though the prediction is correct only for 48 cases.

Second, although the results from our method record relatively lower score than that of the supervised baseline model (DL-BiLSTM+ELMo model) proposed in \newcite{shirani2019learning}, 
we find that it generates quite meaningful emphasis selections.
\begin{table}[!t]
\centering
\scriptsize
\label{tab1}
\sisetup{detect-weight,mode=text}
\renewrobustcmd{\bfseries}{\fontseries{b}\selectfont}
\renewrobustcmd{\boldmath}{}
\newrobustcmd{\B}{\bfseries}
\begin{tabular}{c c c } \hline
    \B Models & \B S1: ``Beauty is not in the face ; beauty is a light in the heart . '' & \B S2: ``The bird a nest , the spider a web , man friendship . '' \\ \hline
    Gold & heart(1), Beauty(2), light(3), in(4), the(4) & friendship(1), bird(2), nest(2), man(2) \\
    Ours & light(1), face(2), Beauty(3), beauty(4) & friendship(1), web(2), a(3), nest(4) \\
    M1 / M2 / M3 / M4 / R & 0.0 / 0.0 / 0.6667 / 0.5 / 0.2917 & 1.0 / 0.5 / 0.3333 / 0.5 / 0.5833 \\ \hline
\end{tabular}
\caption{Top-$4$ emphasis frequency words in two sentences from gold and our model.
Gold: from gold $e\_freq$ values, Ours: DistilBERT-base-uncased's best configuration. Words are sorted based on $e\_freq$ values and ranks are in parentheses. M1-M4, R: sentence-wise $Match_1$-$Match_4$ score and ranking score for the corresponding sentence.}
\end{table}
In Table 3, the DistilBERT-base-uncased model selects \emph{nest, a, web, friendship} to emphasize in S2, which results in \emph{``The bird a \textbf{nest} , the spider \textbf{a web} , man \textbf{friendship} . ''}.
Instead, the gold generates \emph{``The \textbf{bird} a \textbf{nest} , the spider a web , \textbf{man} \textbf{friendship} . ''}, and it seems that our model's result is also valuable.

Third, the DistilBERT models achieve high performance despite its small number of model parameters.
We conjecture that the distillation techniques applied to build the DistilBERT model function as an ensemble of the attention heads from its parent model.
Besides, uncased models show better performance than cased models.
We hypothesize that preserving a word's capital letters is meaningful when selecting proper words to be emphasized.

Lastly, ensembling several PLMs is certainly beneficial.
We ensemble the top-5 models by averaging over their \(e\_freq\) values and achieve the 0.6898 ranking score, which is significantly higher than those of individual PLMs.

\begin{figure}[!h]
    \begin{minipage}[c]{.15\linewidth}
    \end{minipage}
    \hfill
    \begin{minipage}[c]{0.3\linewidth}
        \centering
        \includegraphics[width=\linewidth]{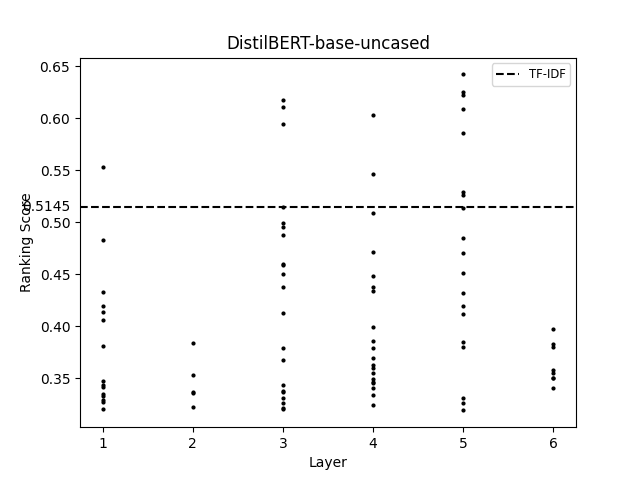}
        \subcaption{DistilBERT-base-uncased}
    \end{minipage}
    \hfill
    \begin{minipage}[c]{.05\linewidth}
    \end{minipage}
    \hfill
    \begin{minipage}[c]{0.3\linewidth}
        \centering
        \includegraphics[width=\linewidth]{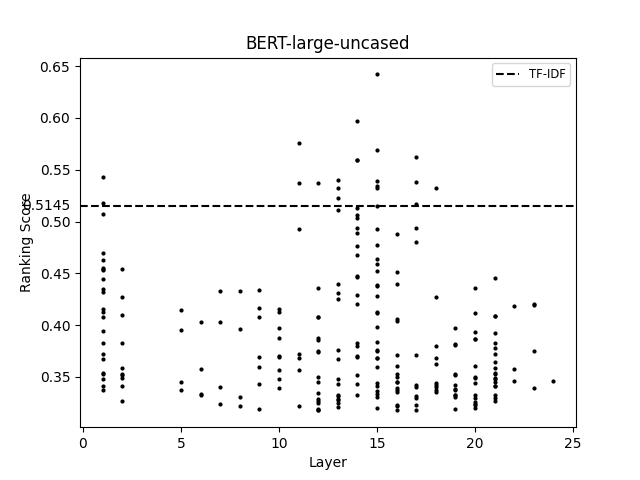}
        \subcaption{BERT-large-uncased}
    \end{minipage}
    \hfill
    \begin{minipage}[c]{.15\linewidth}
    \end{minipage}
    \caption{Layer-wise ranking score of DistilBERT-base-uncased and BERT-large-uncased models.
  Each dot represents a configuration which records the ranking score above that of the random baseline.
  Dotted lines correspond to the ranking score of TF-IDF baseline.}
\end{figure}
For further analysis, we investigate how many attention heads are capable of selecting proper words to emphasize.
In the case of the top-2 models (DistilBERT-base-uncased and BERT-large-uncased), we probe the layer-wise ranking scores of individual attention heads.
We find that there exist attention heads specialized for word emphasis (ones recording high ranking scores).
For both cases, it seems that there is a gap between topmost attention heads and the others.
For instance, the configuration with secondary ranking score reports 0.5975, which is 0.0445 lower than the score of top-$1$ in the BERT-large-uncased model.
\comment{This finding coincides with one reported by \newcite{voita2019analyzing}, where only a few heads in a PLM are effective in machine translation. Likewise, we observe that a few specialized heads are effective in capturing important words.}

\comment{
\subsection{Ablation Study}

\begin{table}[!h]
\centering
\scriptsize
\label{tab2}
\sisetup{detect-weight,mode=text}
\renewrobustcmd{\bfseries}{\fontseries{b}\selectfont}
\renewrobustcmd{\boldmath}{}
\newrobustcmd{\B}{\bfseries}
\begin{tabular}{l c c c c}
\B Model             & \B Full & \B No-diag & \B No-special & \B Only-diag \\
\hline
BERT-base-uncased & 0.6366 & 0.6336 & 0.6397 & 0.5833 \\
BERT-large-uncased & 0.6420 & 0.6415 & 0.6398 & 0.5184 \\
DistilBERT-base-cased & 0.6302 & 0.6246 & 0.6293 & 0.4949 \\
DistilBERT-base-uncased & 0.6420 & 0.6342 & 0.6323 & 0.5808 \\
DistilBERT-base-multilingual & 0.6344 & 0.6306 & 0.6290 & 0.5853 \\
\end{tabular}
\caption{Ablation study. \textbf{Full}: utilizing full attention map including special tokens. \textbf{No-diag}: removing self-attention values from the attention map. \textbf{No-special}: excluding special tokens. \textbf{Only-diag}: only using self-attention values.}
\end{table}

In ablation study, we probe how each component of attention distributions contributes to the performance.
In most cases, except for the BERT-base-uncased model, utilizing special tokens is valuable.
The diagonal component of the attention map, which is self-attention values, is also essential for a better ranking score.
Interestingly, performance degradation occurring when using only self-attention values are quite different among models.
We analyze that the performance drop comes from corresponding models' 
(BERT-large-uncased and DistilBERT-base-cased) tendency to attend to a period or the word itself.
}

\section{Conclusion}
We have proposed a zero-shot emphasis selection method, focusing on investigating whether pre-trained language models contain enough knowledge to select proper words to be emphasized.
We have found that some PLMs report comparable performance, confirming that some specialized attention heads of PLMs have ability to detect meaningful words.

\section*{Acknowledgements}
We would like to thank  Jihun Choi and the anonymous reviewers for their thoughtful
and valuable comments. This work was supported by the National Research Foundation of Korea
(NRF) grant funded by the Korea government (MSIT) (NRF2016M3C4A7952587).

\bibliographystyle{coling}
\bibliography{semeval2020}

\begin{thebibliography}{}

\bibitem[\protect\citename{Clark \bgroup et al.\egroup }2019]{clark2019does}
Kevin Clark, Urvashi Khandelwal, Omer Levy, and Christopher~D Manning.
\newblock 2019.
\newblock What does bert look at? an analysis of bert’s attention.
\newblock In {\em Proceedings of the 2019 ACL Workshop BlackboxNLP: Analyzing
  and Interpreting Neural Networks for NLP}, pages 276--286.

\bibitem[\protect\citename{Conneau and Lample}2019]{conneau2019cross}
Alexis Conneau and Guillaume Lample.
\newblock 2019.
\newblock Cross-lingual language model pretraining.
\newblock In {\em Advances in Neural Information Processing Systems}, pages
  7057--7067.

\bibitem[\protect\citename{Devlin \bgroup et al.\egroup }2019]{devlin2019bert}
Jacob Devlin, Ming-Wei Chang, Kenton Lee, and Kristina Toutanova.
\newblock 2019.
\newblock Bert: Pre-training of deep bidirectional transformers for language
  understanding.
\newblock In {\em Proceedings of the 2019 Conference of the North American
  Chapter of the Association for Computational Linguistics: Human Language
  Technologies, Volume 1 (Long and Short Papers)}, pages 4171--4186.

\bibitem[\protect\citename{Goldberg}2019]{goldberg2019assessing}
Yoav Goldberg.
\newblock 2019.
\newblock Assessing bert's syntactic abilities.
\newblock {\em arXiv preprint arXiv:1901.05287}.

\bibitem[\protect\citename{Jones}1972]{jones1972statistical}
Karen~Sparck Jones.
\newblock 1972.
\newblock A statistical interpretation of term specificity and its application
  in retrieval.
\newblock {\em Journal of documentation}.

\bibitem[\protect\citename{Kim \bgroup et al.\egroup }2020]{Kim2020Are}
Taeuk Kim, Jihun Choi, Daniel Edmiston, and Sang goo Lee.
\newblock 2020.
\newblock Are pre-trained language models aware of phrases? simple but strong
  baselines for grammar induction.
\newblock In {\em International Conference on Learning Representations}.

\bibitem[\protect\citename{Kovaleva \bgroup et al.\egroup
  }2019]{kovaleva2019revealing}
Olga Kovaleva, Alexey Romanov, Anna Rogers, and Anna Rumshisky.
\newblock 2019.
\newblock Revealing the dark secrets of bert.
\newblock In {\em Proceedings of the 2019 Conference on Empirical Methods in
  Natural Language Processing and the 9th International Joint Conference on
  Natural Language Processing (EMNLP-IJCNLP)}, pages 4356--4365.

\bibitem[\protect\citename{Liu \bgroup et al.\egroup }2019]{liu2019roberta}
Yinhan Liu, Myle Ott, Naman Goyal, Jingfei Du, Mandar Joshi, Danqi Chen, Omer
  Levy, Mike Lewis, Luke Zettlemoyer, and Veselin Stoyanov.
\newblock 2019.
\newblock Roberta: A robustly optimized bert pretraining approach.
\newblock {\em arXiv preprint arXiv:1907.11692}.

\bibitem[\protect\citename{Radford \bgroup et al.\egroup
  }2018]{radford2018improving}
Alec Radford, Karthik Narasimhan, Tim Salimans, and Ilya Sutskever.
\newblock 2018.
\newblock Improving language understanding by generative pre-training.
\newblock {\em URL https://s3-us-west-2. amazonaws.
  com/openai-assets/researchcovers/languageunsupervised/language understanding
  paper. pdf}.

\bibitem[\protect\citename{Rogers \bgroup et al.\egroup
  }2020]{rogers2020primer}
Anna Rogers, Olga Kovaleva, and Anna Rumshisky.
\newblock 2020.
\newblock A primer in bertology: What we know about how bert works.
\newblock {\em arXiv preprint arXiv:2002.12327}.

\bibitem[\protect\citename{Sanh \bgroup et al.\egroup
  }2019]{sanh2019distilbert}
Victor Sanh, Lysandre Debut, Julien Chaumond, and Thomas Wolf.
\newblock 2019.
\newblock Distilbert, a distilled version of bert: smaller, faster, cheaper and
  lighter.
\newblock {\em arXiv preprint arXiv:1910.01108}.

\bibitem[\protect\citename{Shirani \bgroup et al.\egroup
  }2019]{shirani2019learning}
Amirreza Shirani, Franck Dernoncourt, Paul Asente, Nedim Lipka, Seokhwan Kim,
  Jose Echevarria, and Thamar Solorio.
\newblock 2019.
\newblock Learning emphasis selection for written text in visual media from
  crowd-sourced label distributions.
\newblock In {\em Proceedings of the 57th Annual Meeting of the Association for
  Computational Linguistics}, pages 1167--1172.

\bibitem[\protect\citename{Shirani \bgroup et al.\egroup
  }2020]{shirani2020semeval}
Amirreza Shirani, Franck Dernoncourt, Nedim Lipka, Paul Asente, Jose
  Echevarria, and Thamar Solorio.
\newblock 2020.
\newblock Semeval-2020 task 10: Emphasis selection for written text in visual
  media.
\newblock In {\em Proceedings of the 14th International Workshop on Semantic
  Evaluation}.

\bibitem[\protect\citename{Vaswani \bgroup et al.\egroup
  }2017]{vaswani2017attention}
Ashish Vaswani, Noam Shazeer, Niki Parmar, Jakob Uszkoreit, Llion Jones,
  Aidan~N Gomez, {\L}ukasz Kaiser, and Illia Polosukhin.
\newblock 2017.
\newblock Attention is all you need.
\newblock In {\em Advances in neural information processing systems}, pages
  5998--6008.

\bibitem[\protect\citename{Vig}2019]{vig2019visualizing}
Jesse Vig.
\newblock 2019.
\newblock Visualizing attention in transformer-based language representation
  models.
\newblock {\em arXiv preprint arXiv:1904.02679}.

\bibitem[\protect\citename{Yang \bgroup et al.\egroup }2019]{yang2019xlnet}
Zhilin Yang, Zihang Dai, Yiming Yang, Jaime Carbonell, Russ~R Salakhutdinov,
  and Quoc~V Le.
\newblock 2019.
\newblock Xlnet: Generalized autoregressive pretraining for language
  understanding.
\newblock In {\em Advances in neural information processing systems}, pages
  5754--5764.

\end{thebibliography}

\end{document}